\documentclass[letterpaper, conference]{IEEEtran}
\IEEEoverridecommandlockouts
\usepackage{cite}
\usepackage{amsmath,amssymb,amsfonts}
\usepackage{algorithmic}
\usepackage[ruled]{algorithm2e} % For algorithms
\usepackage{graphicx}
\usepackage{textcomp}
\usepackage{xcolor}

\usepackage{float}
\usepackage[caption = false]{subfig}

\def\BibTeX{{\rm B\kern-.05em{\sc i\kern-.025em b}\kern-.08em
    T\kern-.1667em\lower.7ex\hbox{E}\kern-.125emX}}
\begin{document}

\title{Implementing a foveal-pit inspired filter in a Spiking Convolutional Neural Network: a preliminary study}
% {\footnotesize \textsuperscript{*}Note: Sub-titles are not captured in Xplore and
% should not be used}

\author{\IEEEauthorblockN{Shriya T.P. Gupta}
\IEEEauthorblockA{\textit{Computer Science \& Information Systems} \\
\textit{BITS Pilani, Goa Campus}\\
Goa, India \\
shriyatp99@gmail.com}
\and
\IEEEauthorblockN{Basabdatta Sen Bhattacharya}
\IEEEauthorblockA{\textit{Computer Science \& Information Systems} \\
\textit{BITS Pilani, Goa Campus}\\
Goa, India \\
basabdattab@goa.bits-pilani.ac.in}
}

\maketitle

\begin{abstract}
We have presented a Spiking Convolutional Neural Network (SCNN) that incorporates retinal foveal-pit inspired Difference of Gaussian filters and rank-order encoding. The model is trained using a variant of the backpropagation algorithm adapted to work with spiking neurons, as implemented in the Nengo library. We have evaluated the performance of our model on two publicly available datasets~---~one for digit recognition task, and the other for vehicle recognition task. The network has achieved up to 90\% accuracy, where loss is calculated using the cross-entropy function. This is an improvement over around 57\% accuracy obtained with the alternate approach of performing the classification without any kind of neural filtering. Overall, our proof-of-concept study indicates that introducing biologically plausible filtering in existing SCNN architecture will work well with noisy input images such as those in our vehicle recognition task. Based on our results, we plan to enhance our SCNN by integrating lateral inhibition-based redundancy reduction prior to rank-ordering, which will further improve the classification accuracy by the network.
\end{abstract}


\section{Introduction}
\label{sec:1}
We have presented a Spiking Convolutional Neural Network (SCNN) that is inspired by two previous works viz. Diehl et al~\cite{diehl2015unsupervised} and Kheradpisheh et al~\cite{kheradpisheh2018stdp}. Both these works have used Leaky Integrate and Fire model of a spiking neuron in a SCNN. A novelty of the SCNN proposed in~\cite{kheradpisheh2018stdp} is the use of Difference of Gaussian (DoG) filters, inspired by the receptive field structure in the primate visual pathway. Furthermore, the preferences of the spiking neurons in their SCNN change progressively, thus capturing different salient features of the input image. Spiking Neural Networks (SNN) are often referred to as the third generation neural networks that hold the potential for sparse and low-power computation~\cite{ghosh2009third}. However, due to the discrete nature of the spiking neuron models, training of SNN using the traditional backpropagation algorithm has been a challenge. A commonly adopted method for training spiking neural networks is the Hebbian learning inspired Spike Time Dependent Plasticity (STDP) algorithm. Thus, both Diehl et al~\cite{diehl2015unsupervised} and Kheradpisheh et al~\cite{kheradpisheh2018stdp} have used STDP-based learning; the former use a two-layer fully connected SCNN, trained using STDP, achieving an accuracy of 95.0\% on the MNIST dataset; the latter use a SCNN with multiple convolutional and pooling layers, achieving an accuracy of up to 98.4\% on the MNIST dataset. 

In a more recent development, researchers have proposed several ways to get round the problem of discontinuity in spiking neurons, allowing them to use backpropagation technique with SNN. For example, Mostafa et al~\cite{mostafa2017supervised} have used backpropagation algorithm in a SCNN by modifying it for the case of non-differentiable spiking neuron response functions, achieving 2.45\% test error rate on the noisy MNIST input dataset. Similarly, Neftci et al~\cite{neftci2017event} have proposed a method known as the event driven random backpropagation that achieves a test classification error of 1.96\% on the MNIST dataset; Lee et al~\cite{lee2018training} have developed a semi-supervised training framework by using a unsupervised STDP rule for pre-training the network, and then using the supervised gradient descent algorithm for fine tuning the model. In another recent development, Hunsberger and Eliasmith~\cite{hunsberger2016training} have proposed a backpropagation training of spiking neurons, which is implemented in the Nengo library~\cite{bekolay2014nengo}. On the contrary, some works adopt a hybrid approach and combine a traditional CNN with a spiking neural network. For instance, Xu et al~\cite{xu2018csnn} propose a hybrid network by using a partial CNN as the initial feature extractor followed by an Inception network made up of LiF neurons that takes the extracted features from the first stage as input. Their SCNN model achieves an accuracy of 88\% on the MNIST dataset using only 500 training samples. They further extend their model with a deeper feature extractor to create Deep CovDenseSNN~\cite{xu2020deep}, a model which generalizes well even for noisy environments and achieve an accuracy of 52\% on the noisy MNIST dataset.

For the SCNN proposed here, we use DoG filters that are inspired by the foveal-pit of the retina. These biologically-inspired filters serve as the feature extractors of our network, which mimic the visual information processing system in the human brain. The DoG functions are implemented as the first filter layer of the network; training the network is done using the  backpropagation algorithm as implemented in the Nengo library~\cite{bekolay2014nengo, hunsberger2016training}.  The foveal-pit model was shown as capable of capturing more than 80\% perceptual information in a static image~\cite{bhattacharya2010biologically}. Inspired by the rank-order encoding in a spatial model of the retina~\cite{rullen2001rate}, the foveal-pit model uses rank-order encoding on information filtered using DoG functions. These filters mimic the receptive fields of the midget and parasol ganglion cells (spiking neurons of the retina) that sub-serve the photo-receptors of the foveal-pit. In our model, we have tested with several combination of the DoG functions that suit the image sizes in our input database. This is done by reconstructing the images after applying the neural filters, and visually inspecting the reconstructed images for quality. The DoG filtering is followed by three layers of convolution and pooling. Finally, there is  a flattening layer, which feeds into the output `dense' (all-to-all connectivity followed by a softmax non-linearity) layer. The output dense layer contains as many neurons as there are classes in our database. It is worth mentioning here that the softmax operation is not implemented through the spiking neurons in our model, rather a softmax function is applied to the outputs of the neurons of the last layer in order to compute the cross entropy loss. 

We have used two datasets in this work viz.\ the MNIST and the Caltech datasets as in~\cite{kheradpisheh2018stdp}. The MNIST database has ten digit images. Thus the dense layer in our proposed SCNN and corresponding to the MNIST database has ten spiking neurons. Two sets of images from the Caltech dataset are used here for vehicle recognition task; the dense layer of our SCNN corresponding to this database has two spiking neurons. The outputs from the neurons of the last layer are expressed in terms of voltages (mV) and are generated using the probe function of the Nengo library~\cite{bekolay2014nengo, nengo}. These outputs are representative of the progressively increasing membrane potentials and the neuron with the highest voltage over a 60 ms simulation time period indicates the class that was selected in a certain epoch. This is followed by a loss computation for the SCNN using a Cross-Entropy function. We have observed a classification accuracy of upto 90\% with our SCNN. When tested without the DoG filtering, the SCNN performance accuracy dropped to around 57\%. However, we have noted that the foveal-pit model, due to its high overlapping DoG functions, needs lateral inhibition mechanism implemented before rank-order encoding. Incorporating lateral inhibition in the neural filters of our proposed SCNN will be taken up as future work. The filtering step is applied to the entire set of training images as a pre-processing step. These filtered images serve as the input to our subsequent classification network. 

In Section~\ref{sec:2}, we present the methodology used in our work. In Section~\ref{sec:3}, we present the results of our proof-of-concept study. Conclusions and the future directions for the work are presented in Sec.~\ref{sec:4}. The code used in this work is open-sourced~\footnotemark.
\footnotetext{https://github.com/shriya999/spikingCNN}
\section{Methodology}
\label{sec:2}
\subsection{Datasets}
\label{sec:21}
The primary aim of this preliminary study is to propose a model for the classification of images in both noisy and noise-free environments. For this, we have used the MNIST dataset~\cite{lecun1998gradient} for the noise-free scenario and a subset of the Caltech dataset~\cite{fei2004learning} for the noisy case. Readers may note that images from the MNIST dataset are clean images as they only contain digits without any associated background. Conversely, images from the Caltech dataset are noisy in the sense that they have corresponding background details as the images are from a real world setting.
\subsubsection{Digit Recognition}
The MNIST digit recognition database is considered as the benchmark for image recognition tasks. In this, each image is of size $28 \times 28$ and represents a digit between 0 to 9. In total, it has 55,000 training images and 10,000 testing images. For each image, there is a predetermined label $l$ which is an integer value representing the numerical value of the digit. We have used the MNIST dataset from the Tensorflow repository as input to our SCNN, for ease of access and to minimize the storage requirements of our model. This repository does not contain the entire set but only a subset of 55,000 training images.

\subsubsection{Vehicle Recognition}
In order to test the robustness of our model on noisy images, we sampled a subset of Caltech 101 image dataset belonging to the motorbike and airplane classes. This subset has roughly 800 images from each class with varying dimensions. For purposes that suit the filter dimension in this work, we have resized the images in this sample dataset to $150 \times 200$ pixels.

\begin{figure*}[thpb] 
\centering
      \includegraphics[scale=0.6]{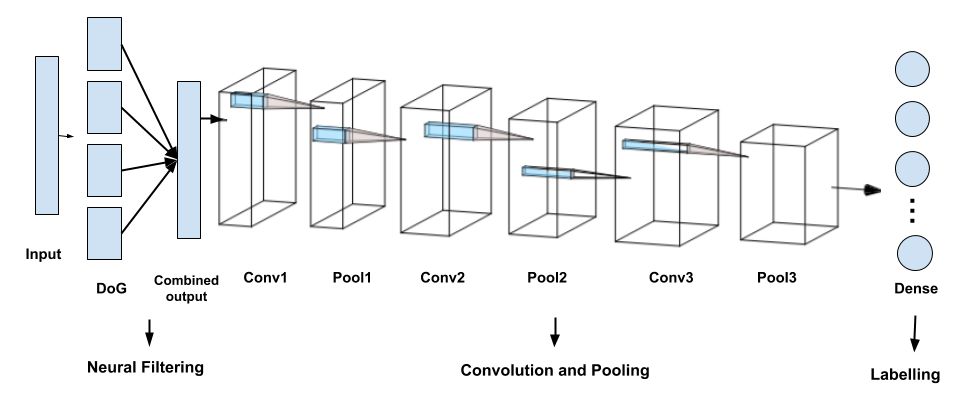}
      \caption{The overall architecture of the proposed Spiking Convolutional Neural Network.}
      \label{CNNnet}
\end{figure*}

The architecture of our proposed SCNN is shown in Fig.~\ref{CNNnet} and is described in the following sections.
\subsection{Retina-inspired filtering and processing}
\label{sec:22}
\subsubsection{Difference of Gaussian Receptive Fields}
The foveal-pit is a circular region of 200 $\mu m$ diameter in the centre of the fovea. Because of its structure, it is the region that is most accessible to incoming light, and has the highest visual acuity in the primate retina. Sensory signals from the photo-receptors pass through layers of retinal neurons of other types before reaching the ganglion cells. The retinal ganglion cells are the only spiking neurons of the primate retina, and it is their axons that transmit the sensory signals from the retina to the higher parts of the brain. There are mainly two types of ganglion cells in the primate retina viz. midget and parasol ganglion cells. Each of these ganglion cell types has centre-surround receptive fields of two types --- on-centre-off-surround and off-centre-on-surround. These are typically modelled using Difference of Gaussian (DoG) functions.

\begin{figure}[!thpb] 
\centering
\subfloat[]{\includegraphics[width = 0.8in]{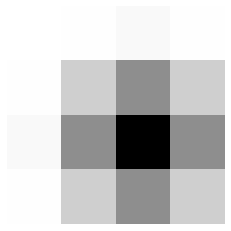}}
\hspace{3em}
\subfloat[]{\includegraphics[width = 0.8in]{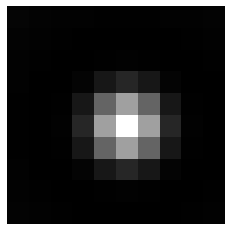}} \\
\subfloat[]{\includegraphics[width = 0.8in]{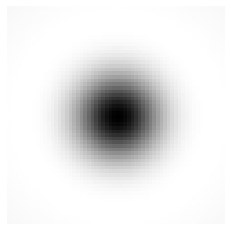}}
\hspace{3em}
\subfloat[]{\includegraphics[width = 0.8in]{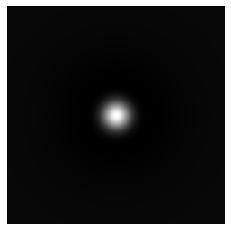}}
\caption{The ganglion cells modelled using DoG functions representing the (a) off-center midget cell (b) on-center midget cell (c) off-center parasol cell and (d) on-center parasol cell.}
\label{ganglion}
\end{figure}

In this work, we have used biologically inspired neural filters as done by~\cite{kheradpisheh2018stdp}, simulating retinal information processing. However, we have modelled the receptive fields of the ganglion cells that sub-serve the retinal foveal-pit as in~\cite{bhattacharya2010biologically}. A sample of the filter types and dimension are shown in Fig.~\ref{ganglion}. The off- and on-center midget cells are modelled with $5 \times 5$ and $11 \times 11$ DoG functions respectively, where the standard deviation of the centres are 0.8 and 1.04; the off- and on-center parasol cells are modelled with $61 \times 61$ and $243 \times 243$ DoG functions respectively, with standard deviation of the centres set to 8 and 10.4. The surround width for midget cells is 6.7 times that of the center, while that for parasol cells is 4.8 times the center width. The input image of size $M \times N$ is convolved with each of these filter types, with stride of 1, resulting in four matrices each of size $M \times N$. The Algorithm~\ref{algo} describes the method to combine the outputs of the convolution operation.
\begin{algorithm}
 \label{algo}
 \caption{Algorithm for combining filtered outputs. The output of filtering with: midget off- and on-centre cells are called moff\_img and mon\_img respectively; parasol off- and on-centre cells are called poff\_img and pon\_img respectively.}
 \begin{algorithmic}[1]
    \STATE combined\_outputs = []
    \FOR{i in train\_set}
    \STATE final\_img = zeros (i.shape[0], i.shape[1])
    \STATE mon\_img = convolve (i, mon\_filt)
    \STATE moff\_img = convolve (i, moff\_filt)
    \STATE pon\_img = convolve (i, poff\_filt)
    \STATE poff\_img = convolve (i, pon\_filt)
    \FOR{j = 0 to j = i.shape[0] }
    \FOR {k = 0 to k = i.shape[1] }
    \STATE final\_img [j][k] = max\{ mon\_img [j][k], moff\_img [j][k], pon\_img [j][k], pon\_img [j][k] \}
    \ENDFOR
    \ENDFOR
    \STATE Append final\_img to combined\_outputs
    \ENDFOR
    \RETURN combined\_outputs
 \end{algorithmic} 
\end{algorithm}
\subsubsection{Rank-order encoding}
Rank-order neural codes were first demonstrated in~\cite{rullen2001rate}, and subsequently used in~\cite{bhattacharya2010biologically} for encoding static visual information with the foveal-pit model. Rank-order encoding assumes that in the time taken by the visual cortex to perceive information from the environment travelling from the retina up the visual pathway, each neuron in the retina can fire no more than one spike; thus, it is the time to spike that forms the basis of encoding information by the retina. The very first retinal spiking neurons (ganglion cells) that fire a spike are thought to carry the most important information about the environment, and so on.

We follow this method here. The co-efficient of convolution matrices described in the previous section are flattened to form a 1-D array of dimension $Q = 4MN$. To implement rank-order encoding, they are sorted in descending order according to their pixel strength. The assumption is that the highest pixel (coefficient) value correspond to the first ganglion cell to have fired, and carries the most weight (importance), and so on. Thus, each of the sorted coefficients are weighted in a progressively decreasing order.

In this work, we retain only the top $M \times N$ rank-order coefficients, which is reshaped to a 2-D array. To enable this conversion from 1-D to 2-D, we need to preserve the spatial information of each coefficient of the convolution operation. Furthermore, to be able to qualitatively test our rank-ordering scheme (see below), we need to preserve information of the exact kind of DoG that was used to obtain each of the rank-ordered coefficients. Thus, we created a set of two dictionaries, $D_1$ and $D_2$~---~the former maps the coefficients $C_i$ to their respective DoG filters $F_i$, while the latter maps each of the coefficients $C_i$ to a tuple $(x_i,y_i)$ representing the x-y co-ordinates of its original location (i.e.\ before the rank-order encoding operation). We thus have $M \times N$ rank-ordered features of the original image, that is used to drive the first convolutional layer.

\textit{Qualitative Validation:} Before we proceed to the next step, we qualitatively validate the outcome of our rank-order encoding using the foveal-pit informed DoG filters. For each coefficient $C_i$ in the rank-ordered image matrix, we accessed its corresponding DoG filter $F_i$ from the dictionary $D_1$ and its spatial coordinates $(x_i,y_i)$ from dictionary $D_2$. Reconstruction methods using these information are exactly as in~\cite{rullen2001rate}. Our foveal-pit based filters in the preprocessing step of the model help in noisy environments as the larger parasol cells capture more of the background information whereas the smaller midget cells are able to pick up the finer details of the image. Hence, using a combination of these ganglion cell inspired filters in our proposed approach helps in successfully eliminating the background noise and to obtain a good accuracy for both noisy and noise-free environments. The results are discussed in Sec.~\ref{sec:3}. 

\subsection{Convolution with Spiking Neurons}
\label{sec:23}
The spiking neurons in the CNN in this work consist of Leaky Integrate and Fire (LIF) neurons as in~\cite{diehl2015unsupervised},~\cite{kheradpisheh2018stdp}, and are not defined here for brevity. A layer of LIF neurons of the same size $M \times N$ of the input image is used to convert the rank-ordered image features to spikes. Each rank-ordered co-efficient is mapped to an input current value that will drive the spatially-corresponding LIF. We now have a $M \times N$ spiking neural layer $\mathcal{S}$.

Convolution is done in three consecutive layers with filter sizes $3 \times 3$. This is indicated in Fig.~\ref{CNNnet}. There are $2^3$, $2^4$ and $2^5$ filters respectively in the first, second and third convolution layers, that define the thickness of the result of convolution, indicated by qualitatively increasing depth of the representative boxes in Fig.~\ref{CNNnet}. As is conventional, a pooling operation follows each convolution operation. In this work, we have used non-linear average pooling with a stride of 2. Average pooling is used as it eliminates some of the redundant information captured by adjacent neurons in the same window that have overlapping inputs.

The output of the spiking neural layer $\mathcal{S}$ is convolved with the $3 \times 3$ filters with a stride of 1 and padding as appropriate so that the size of the output of the first convolutional layer is a $M \times N \times 2^3$. After pooling with stride 2, the size of the output layer will be $M/2 \times N/2 \times 2^3$. The exact sizes of the layers in our work for both datasets are mentioned in Table.~\ref{dimension}.
\begin{table}[!thpb]
\renewcommand{\arraystretch}{1.0}
\caption{Dimensions of each layer in the proposed neural network.}
\label{dimension}
\centering
 \begin{tabular}{|c|c|c|c|} 
 \hline
Layer &MNIST& Caltech & Kernel size\\[0.5ex]
\hline
Conv1 &(28,28,1)& (150,200,1)& 3 \\
Pool1 &(28,28,8)& (150,200,8) & 2 \\
Conv2 &(14,14,8)& (75, 100, 8) & 3 \\
Pool2 &(14,14,16)& (75,100,16) & 2 \\
Conv3 &(7,7,16)& (37,50,16) & 3 \\
Pool3 &(7,7,32)& (37,50,32) & 2 \\
Flatten &(3,3,32)& (18,25,32) &  -- \\
Dense &(1,288)&  (1, 14400) & -- \\
Outputs & (1,10) & (1,2) & --\\
 \hline
\end{tabular}
\end{table}
To put the convolution and pooling operations in perspective, i.e.\ in context to biological visual processing, we may say this:
The visual features extracted during each convolutional filtering layer is a combination of several simple features extracted from previous layers; thus, the image recognition task is carried out in a hierarchical fashion. Through this hierarchy of layers, the preference of `neurons' gradually change from primarily edge detection in the first convolutional layer to more surrounding environmental information in the third convolutional layer. The synaptic connections between the pre-synaptic `neurons' of one layer and the post-synaptic `neurons' of the next layer are modelled as the weights of the network that are adjusted during the training period. We discuss the training and learning in section~\ref{sec:25}. Prior to training, the data will need to be prepared accordingly, and is described in the next section.
\subsection{Loss computation}
\label{sec:24}
The output of the final convolution-pooling operation is converted to a column vector using a flatten operation. This vector is fed with all-to-all connectivity to the last layer of the network that contains $K$ spiking neurons, where $K$ is the number of distinct classes in our classification problem. Thus, the output layer of our network corresponding to the digit recognition task will have $K$ = 10 neurons, while that corresponding to the vehicle recognition task will have $K$ = 2 neurons. The $K \times 1$ output vector $\mathbf{X}: \mathbf{X} \in \mathcal{R}^K$ is converted into a set of probabilities using a softmax function defined as: 
\begin{equation}
\label{eq:f2}
\sigma(x_i) =  \frac{ e^{x_i} } {\sum_{j=1}^{K} e^{x_j} } \forall  i = 1,\cdots,K
\end{equation}
where $x_i \in \mathbf{X}$; we call this softmaxed probability vector as $\mathbf{Y}$. Since the ground truth label for each input image is a single integer, we first convert it into a one-hot encoded label vector $\mathbf{L}$ of dimension $K \times 1$ and representing the ground truth class of each image. Then using the softmaxed probabilities in $\mathbf{Y}$ and the pre-defined and one-hot encoded ground truth vector $\mathbf{L}$, we calculate the cross entropy loss as follows: 
\begin{equation}
\label{eq:f3}
H(\mathbf{L},\mathbf{Y}) = \sum_{i=1}^{K} l_i* \log y_i, 
\end{equation}
where $l_i \in \mathbf{L}$ and $y_i \in \mathbf{Y}$. This loss value is taken as the overall loss of the network which is minimised during each iteration of the training period.
\subsection{Training the network}
\label{sec:25}
For training our network, we use a spiking approximation of the backpropagation algorithm similar to the one used in artificial neural networks. However, because spiking neurons are not differentiable, we use an approximation of LIF neurons that can be differentiated as proposed by  \cite{hunsberger2016training}, with the original firing rate of LIF neurons defined as:
\begin{equation}
\label{eq:f4}
r(j) = \frac{1}{q(j)} 
\end{equation}
\begin{equation}
\label{eq:f5}
q(j) = \tau_{ref} + \tau_{RC} * \log (1 + \frac{v_{th}}{\rho(j-v_{th})} )
\end{equation}
where $j$ is the input current to the neuron, $\tau_{ref}$ is the refractory period duration,  $\tau_{RC}$ is the time constant and $\rho(x) = max(x,0)$. On differentiating this function w.r.t the input current, we get:
\begin{equation}
\label{eq:f6}
r'(j) = \frac{1}{q(j)^2} * \frac{v_{th}  \tau_{RC}}{ 1 + \frac{v_{th}}{\rho(j-v_{th})}  } * \rho'(j-v_{th})
\end{equation}
But when the input current becomes equal to $v_{th}$, the derivative $r'(j)$ becomes undefined as the denominator has the term $\rho(j-v_{th})$, which becomes zero. Thus, instead of a simple maximum function, a softened LIF response function is used by replacing $\rho(x)$ with  $\log(1+e^x)$ which is still defined when $x$ becomes zero. So, during the training period, we replace the LIF spiking neurons with neurons having this softened response function. Then using the original gradient descent algorithm \cite{lecun1988theoretical}, the weights of our network are adjusted proportional to the net error of the model as follows:
\begin{equation}
\label{eq:f7}
\delta W \propto - \frac{\partial e}{\partial W}
\end{equation}
\begin{equation}
\label{eq:f8}
\delta w_{ji} = - \epsilon \frac{\partial e}{\partial a_j} * \frac{\partial a_j}{\partial net_j} * \frac{\partial net_j}{\partial w_{ji}}
\end{equation}
where $w$ is the synaptic connection weight from the $i^{th}$ presynaptic neuron to the $j^{th}$ postsynaptic neuron, $a_j$ is the activation function of the network which is the firing rate r of the $j^{th}$ LIF neuron in our model and $net_j$ is the net input to the $j^{th}$ neuron. In Eq. \ref{eq:f8}, we keep the first and third differential terms same as the definitions used in the backpropagation algorithm. However, using the above mentioned approximation, we set the second term to:
\begin{equation}
\label{eq:f9}
\frac{\partial a_j}{\partial net_j} = r'(net_j)
\end{equation}
These gradients are then backpropagated from the last output layer to the input layer during each epoch and the synaptic weights of the network are adjusted accordingly. This iterative procedure is carried out for up to 10 epochs for each input image, which is the ideal amount of time the network requires to converge, beyond which it starts to overfit. For the digit recognition task, we perform the training for a set of 55,000 input images and for the vehicle recognition task, this procedure is carried out for a corpus of 1400 training images.
\subsection{Generating predictions}
\label{sec:26}
During the testing phase, we evaluate the performance of our network on 10,000 images from the testing corpora for the digit recognition task and on 200 testing images for the vehicle recognition task. We propagate each input image through the trained network for a total of 30 timesteps for the MNIST dataset and for a total of 60 timesteps for the Caltech dataset and the output of the last layer of our network is a $K$x1 vector, where $K$ is the number of classes. Hence, the predicted label for the input image is assigned as the index of the neuron corresponding to the maximum spike count amongst all the $K$ values. This label is then compared with the ground truth label of the input image to determine the classification accuracy of our network.
\section{Experimental Methods and Results}
\label{sec:3}
The SCNN and its various layers were coded using the Nengo library \cite{bekolay2014nengo} and simulated on GPU accessed via Google Colaboratory \cite{colab}. During the testing phase, we ran our SCNN for a total of 60 timesteps, in order to get an accurate measure of the spiking neurons' outputs over time. However, the labels for each image were assigned using only the spike count of the last timestep.
\subsubsection{Training for Digit Recognition}
\begin{figure}[thpb] 
\centering
\subfloat[]{\includegraphics[width = 1in]{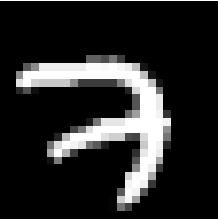}}
\hspace{1em}
\subfloat[]{\includegraphics[width = 1in]{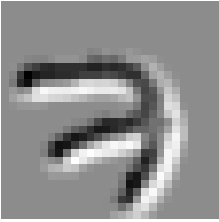}}
\hspace{1em}
\subfloat[]{\includegraphics[width = 1in]{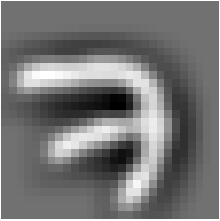}}\\
%\subfloat[]{\includegraphics[width = 1in]{images/7_poff.png}}
%\hspace{1em}
%\subfloat[]{\includegraphics[width = 1in]{images/7_pon.png}}
\caption{(a) The original image of the digit 7. The outputs of the convolution with (b) off-center midget cell (c) on-center midget cell.}
\label{digit_dog}
\end{figure}
For the digit recognition task, we ran our network on the MNIST dataset. As mentioned in Sec.~\ref{sec:22}, the qualitative validation of the foveal-pit DoG filtering is shown in Fig.~\ref{digit_dog}. We observed that because the parasol cell receptive field dimensions (63 and 253) are much larger than the MNIST image size, they do not add any value to the information capture (not shown here). Thus, we do not use these DoG filters for the MNIST dataset. Furthermore, using both midget filters together do not add value to the learning task. Thus, we test the training of the network for two cases viz.\ using rank-ordered output of filtering with on-centre and off-centre midget cells independently. 
\begin{table}[!thpb]
\renewcommand{\arraystretch}{1.0}
\caption{Accuracies (\%) for the digit recognition task}
\label{mnist}
\centering
 \begin{tabular}{|c|c|c|c|} 
 \hline
Neural filter & 3 epochs  &  6 epochs  & 10 epochs \\[0.5ex]
\hline
off-center midget cell & 98.25 & 99.00 & 96.50 \\
on-center midget cel & 97.75 & 98.25 & 96.25 \\
 \hline
\end{tabular}
\end{table}
The results are summarized in Table \ref{mnist}. The best results are achieved with off-center midget cells as these have the smallest dimensions, and are able to capture the finer details of the data. Further, the best results of 99.00\% are achieved for six epochs which is the ideal amount of time required for the network to converge. This accuracy is achieved for the MNIST testing set. On running longer, the network tends to overfit, i.e. learn very specific features related to the training data and so perform poorly on the testing data.
\begin{figure}[thpb] 
%\centering
\subfloat[]{\includegraphics[scale=0.78]{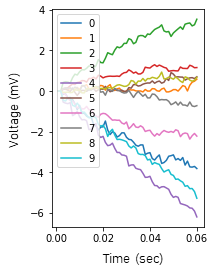}}
\hspace{0.05em}
\subfloat[]{\includegraphics[scale=0.78]{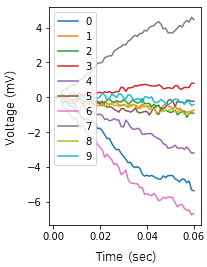}}
\caption{Outputs of the final layer neurons for (a) digit 2 and (b) digit 7. The timestep of simulation is 1 ms and total time of simulation is 60ms.}
\label{mnist plots}
\end{figure}

In the plots of Fig. \ref{mnist plots}, we can see the outputs of the final layer neurons during the testing phase. These outputs are represented in terms of voltages (mV) and are generated using the probe function of the Nengo library~\cite{bekolay2014nengo, nengo}. For an input image of the digit two, the output from the second neuron is the highest and hence this image is classified as two by the network. So, the losing neurons are the ones with negative voltages i.e. lowest values. Similarly, we can see that for the digit seven, the highest output is for the seventh neuron as the input image corresponds to the number seven. 
\begin{table}[thpb]
\renewcommand{\arraystretch}{1.0}
\caption{Comparison of results on the Digit Recognition task}
\label{comparison}
\centering
 \begin{tabular}{|c|c|c|c|c|} 
 \hline
Model & Learning Rule  & Accuracy (\%) \\[0.5ex]
\hline
Two layer SNN~\cite{diehl2015unsupervised} & STDP & 95.00 \\
Convolutional SNN~\cite{diehl2015fast} & Backpropagation& \textbf{99.10} \\
Spiking Deep CNN~\cite{kheradpisheh2018stdp} & STDP & 98.40 \\
Proposed SCNN & Backpropagation & 99.00 \\
 \hline
\end{tabular}
\end{table}

Additionally, we compare our proposed methodology to the existing techniques on the MNIST dataset in Table \ref{comparison}. As seen here, our SCNN approach with foveal-pit inspired filters outperforms the two layer SNN model~\cite{diehl2015unsupervised}, that was trained in an unsupervised manner using the STDP algorithm. It also performs better than the Spiking deep CNN proposed by Kheradphisheh et al~\cite{kheradpisheh2018stdp}, similarly trained with STDP. Moreover, our approach has a comparable performance to the state-of-the-art Convolutional SNN~\cite{diehl2015fast}, which was trained in a supervised manner using a backpropagation algorithm.

\subsubsection{Training for Vehicle Recognition}
\begin{figure}[thpb] 
\centering
\subfloat[]{\includegraphics[width = 1in]{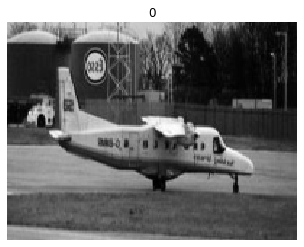}}
\hspace{1em}
\subfloat[]{\includegraphics[width = 1in]{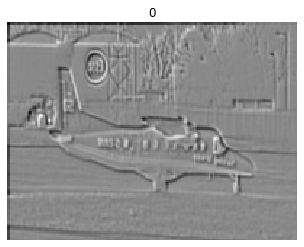}}
\hspace{1em}
\subfloat[]{\includegraphics[width = 1in]{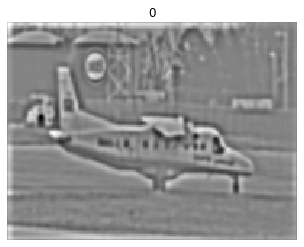}}\\
\subfloat[]{\includegraphics[width = 1in]{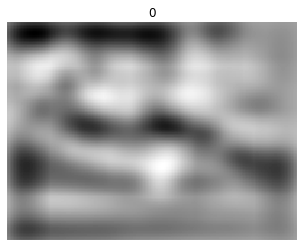}}
\hspace{1em}
\subfloat[]{\includegraphics[width = 1in]{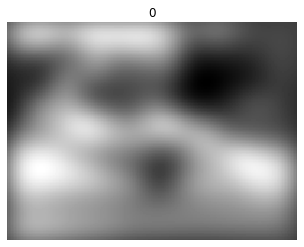}} 
\caption{(a) The original image of an airplane. The outputs of the convolution with (b) off-center midget cell (c) on-center midget cell (d) off-center parasol cell and (e) on-center parasol cell.}
\label{mbike_dog}
\end{figure}

For the vehicle recognition task, images in the dataset have a larger (compared to that of MNIST) dimension of $150 \times 200$ pixels and are noisy. The outputs of convolution with all four DoG filters for an airplane image along with the background noise is shown in Fig.~\ref{mbike_dog}. As seen from the images, the midget cells pick up the finer details of the airplane, whereas the parasol cells capture low resolution contents of the background. So, in each case, we combine a subset of the convolved images using the Algorithm \ref{algo} outlined in Sec.~\ref{sec:2}. 

\begin{figure}[thpb] 
\centering
\subfloat[]{\includegraphics[width = 1in]{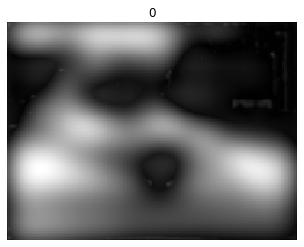}}
\hspace{1em}
\subfloat[]{\includegraphics[width = 1in]{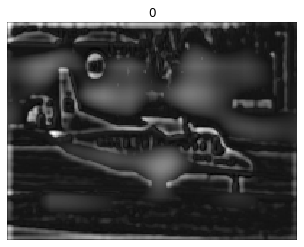}}
\hspace{1em}
\subfloat[]{\includegraphics[width = 1in]{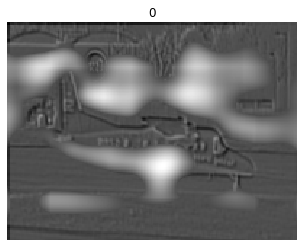}}\\
\subfloat[]{\includegraphics[width = 1in]{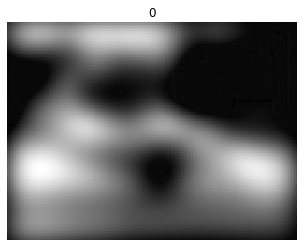}}
\hspace{1em}
\subfloat[]{\includegraphics[width = 1in]{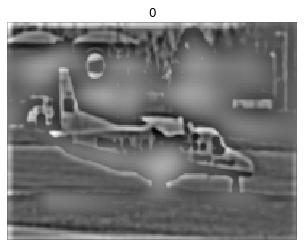}} 
\hspace{1em}
\subfloat[]{\includegraphics[width = 1in]{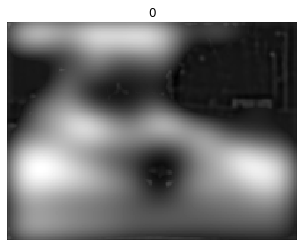}} 
\caption{The final images after combining the filtered outputs of the convolution with (a) all four types of DoG cells (b) off-center midget cell, on-center midget cell, and off-center parasol cell (c) off-center midget cell and off-center parasol cell (d) off-center midget cell and on-center parasol cell (e) on-center midget cell and off-center parasol cell (f) on-center midget cell and on-center parasol cell.}
\label{combo_dog}
\end{figure}

In order to determine the most optimal set of DoG filters, we plot the outputs of the various combinations of filtered outputs using Algorithm \ref{algo} which is shown in Fig.~\ref{combo_dog}. The most optimal combination corresponds to Fig.~\ref{combo_dog} (b). This is because the midget cells capture majority of the intricate details, while the off-center parasol cells capture the background details. Again, we observe that the dimensions of the parasol on-centre filter dimensions are much larger than that of the original image, and therefore do not contribute to capturing any useful information.

\begin{figure}[thpb] 
\centering
\subfloat[]{\includegraphics[width = 1in]{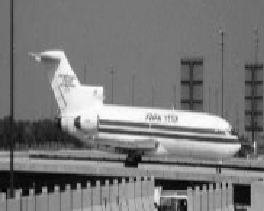}}
\hspace{1em}
\subfloat[]{\includegraphics[width = 1in]{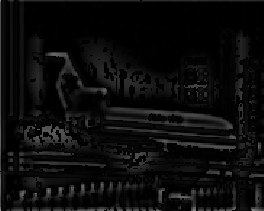}}
\hspace{1em}
\subfloat[]{\includegraphics[width = 1in]{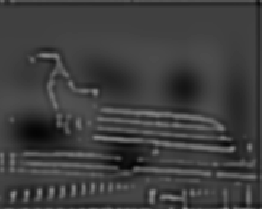}}\\
\subfloat[]{\includegraphics[width = 1in]{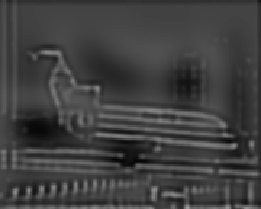}}
\hspace{1em}
\subfloat[]{\includegraphics[width = 1in]{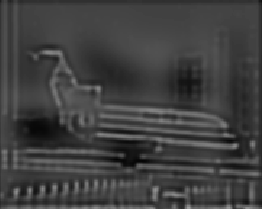}} 
\hspace{1em}
\subfloat[]{\includegraphics[width = 1in]{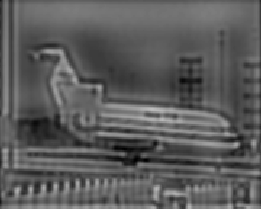}} 
\caption{The output images after the qualitative analysis. (a) The original image of the airplane (b) the rank ordered output of the airplane  (c) the reconstructed image using only 10\% of the coefficients (d) using only 20\% of the coefficients (e) using only 30\% of the coefficients and (f) using all the coefficients for the reconstruction.}
\label{ro_plane}
\end{figure}

For a qualitative understanding of the effect of the neural filtering on noisy Caltech dataset~\cite{fei2004learning}, we show the case where off- and on-center midget cells and the off-center parasol cell DoGs are used for filtering the input image. For the image of an airplane, the outputs of the rank order coding as well as the reconstruction are shown in Fig.~\ref{ro_plane}. 
We note that the image is recognisable even when  we have used only 10\% of the coefficients. Using a larger number of coefficients for reconstruction increases the clarity of the reconstructed image. However, this is only up to a certain point, after which no more information can be retrieved. This is as in rank-order decoding~\cite{rullen2001rate}. Furthermore, the rank-ordered image in Fig.~\ref{ro_plane}(b) indicates the detailed capture of all edges in the image. The absence of lighter portions of the background may be attributed to the lack of on-centre parasol cell filtering. While this image may be speculated to give a high objective measure on perceptual quality (see~\cite{bhattacharya2010biologically} for details), and yet, the accuracy score with cross-entropy is lower than the case of filtering with just midget off-centre DoG functions as elucidated in the following text. This is an interesting observation in our proof-of-concept study, and raises open questions on the suitability of purely data-based algorithms for evaluating classification accuracy of real scenes.

\begin{table}[thpb]
\renewcommand{\arraystretch}{1.0}
\caption{Accuracies (\%) for the vehicle recognition task}
\label{caltech}
\centering
 \begin{tabular}{|c|c|c|c|c|} 
 \hline
Neural filter & 3 epochs  &  5 epochs  & 7 epochs \\[0.5ex]
\hline
without any filtering & 57.50 & 52.50 & 47.50 \\
only off-center midget cells & 90.00 & 52.50 & 67.50 \\
midget-off, midget-on, parasol-off & 60.00 & 52.50 & 52.50\\
 \hline
\end{tabular}
\end{table}
The results of our proposed network for the airplane/motorbike dataset is summarized in Table~\ref{caltech}. In order to define a baseline for this task, we ran the network sans foveal-pit DoG filtering, which achieved an accuracy of up to 57.5\%. The performance is quite low for this case as we have a limited number of images for each class. On convolving the image with the off-center midget cell as the DoG filter, the performance improves to a great extent achieving a highest accuracy of 90.00\% when the model is run for three iterations. The performance also improves slightly leading to an accuracy of up to 60.00\% when we combine the outputs of the off-center midget cell, the on-center midget cell, and the off-center parasol cell as shown in Fig.~\ref{combo_dog}(b). However, the best results are achieved for the case of off-center midget cell mainly because they have the smallest dimensions and so are able to capture the finer details of the image. 

\begin{figure}[thpb] 
\centering
\subfloat[]{\includegraphics[width=2.8in]{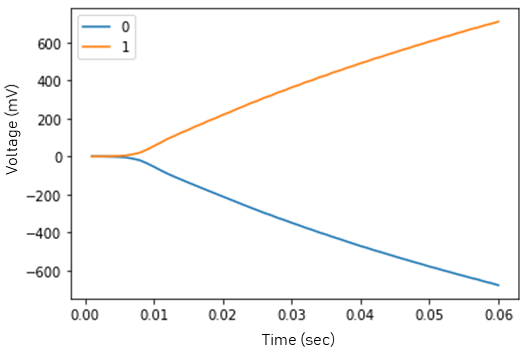}}
\\
\subfloat[]{\includegraphics[width=2.8in]{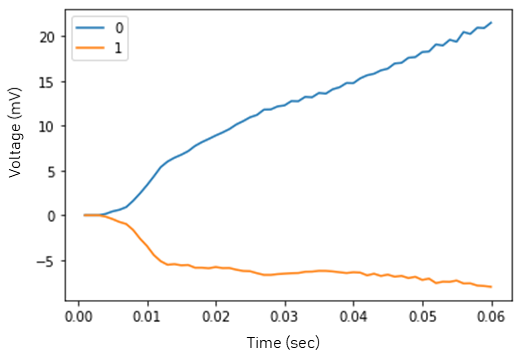}}
\caption{Outputs of final layer neurons for the (a) airplane and (b) motorbike image. The $0$ in the figure legend is for the output of neuron $0$, which corresponds to the motorbike category; The $1$ in the figure legend is for the output of neuron $1$, which corresponds to the airplane category. The timestep of simulation is 1 ms and total time of simulation is 60ms.}
\label{caltech plots}
\end{figure}

The network is trained such that the neuron $0$ will represent the motorbike category, and the neuron $1$ will represent the airplane category. In the plots of Fig.~\ref{caltech plots}, we can see the outputs from the final layer neurons during the testing phase, obtained using the probe function of the Nengo library~\cite{bekolay2014nengo, nengo}. For an input image of an airplane, the output from the neuron $1$ of the output layer is the highest and hence this image is classified correctly by the network. For the motorbike image, the highest output is for the neuron $0$, once again classifying correctly the input image.

\section{Conclusion and Future Work}
\label{sec:4}
In this paper, we present a Spiking Convolutional Neural Network (SCNN) with retinal foveal-pit inspired Difference of Gaussian (DoG) filters to enable the classification of images in both a noise-free and a noisy environment. Before feeding to convolution-pooling layers, the output of the DoG filters are encoded in rank-order and cropped to retain the most salient information in the image. We have explored the qualitative effect of the filtering on the input image to identify the optimal combination of DoG filters for each of the datasets used in our experimental analyses. 

To train our network, we have used a supervised training algorithm that is a modified version of the traditional backpropagation algorithm used in artificial deep learning networks. To evaluate the performance of our model, we experimented with two kinds of datasets, one for digit recognition and another for vehicle recognition. Our model achieves promising results on both of these datasets, achieving an accuracy of up to 99.00\% on the MNIST dataset which is comparable to existing SCNN models. In addition, our model performs well on noisy images from the airplane/motorbike subset of the Caltech dataset, successfully differentiating between the surrounding and the object of interest; this in spite the limited number of images available for training our network. For this binary classification task of vehicle recognition, our model achieves the best accuracy of up to 90.00\%; this is an improvement on around 57\% over the alternate approach of performing the classification without any kind of neural filtering. It is worth mentioning here that the baseline for the vehicle recognition task is not pre-defined, i.e.\ it is generated during the course of this work, and with fewer data compared to that for MNIST. Our results imply the importance of the bio-inspired neural filters in redundancy reduction of input images, and discarding irrelevant background information.

A reconstruction of the rank-ordered image is made following the same procedure as proposed in~\cite{rullen2001rate}. However, because of the comparatively high density of receptive fields in the Foveal-pit model, it is shown to work only with lateral-inhibition based Filter-overlap Correction Algorithm (FoCal) for efficient information capture with sparse data~\cite{bhattacharya2009focal}. However, we could not implement the FoCal due to computational constraints, and therefore our reconstructed images are qualitatively worse than expected; and yet, the accuracy of our proposed SCNN increases for certain combination of filters as demonstrated in Sec.~\ref{sec:3}. Further testing upon implementing the FoCal will be taken up as immediate future work. In addition, unsupervised training using Hebbian learning algorithms will be added to the network for a comparative study with supervised learning using the backpropagation algorithm. Overall, we believe that our preliminary study will contribute to current research in biologically-inspired sparse computational paradigms such as the spiking convolutional neural networks.

\bibliographystyle{./IEEEtran}
\bibliography{./IEEEabrv,./ref}

\end{document}